%% file: camera-ready.tex
\documentclass[conference]{IEEEtran}

\input{preamble}
\begin{document}
\title{
Non-Local Feature Aggregation on Graphs via Latent Fixed Data Structures}

\author{\IEEEauthorblockN
{\bf  Mostafa Rahmani,\  Rasoul Shafipour,\ Ping Li\vspace{0.08in}}
\IEEEauthorblockA{Cognitive Computing Lab \\
Baidu Research\\
10900 NE 8th St. Bellevue, WA 98004, USA \\
\{rahmani.sut,\ rashafipour,\ pingli98\}@gmail.com}

}

\maketitle

\begin{abstract}

In contrast to  image/text data whose order can be used to perform non-local feature aggregation in a straightforward way using the pooling layers, graphs lack the tensor representation and mostly the element-wise max/mean function is utilized to aggregate the locally extracted feature vectors. In this paper, we present a novel approach for global feature aggregation in Graph Neural Networks (GNNs) 
which utilizes a Latent Fixed Data Structure (LFDS) to aggregate the extracted feature vectors. The locally extracted feature vectors are sorted/distributed on the LFDS and a latent neural network (CNN/GNN)  is utilized to perform feature aggregation on the LFDS. The proposed approach is used to design several novel global feature aggregation methods based on the choice of the LFDS. We introduce multiple LFDSs including loop, 3D tensor (image), sequence,  data driven graphs and an  algorithm which sorts/distributes the extracted local feature vectors on the LFDS. 
While the computational complexity of the proposed methods are linear with the order of input graphs, they achieve competitive or better results.
\end{abstract}


\IEEEpeerreviewmaketitle

\section{Introduction}
Deep Neural Networks have shown  superior performance in analyzing data with spatial or temporal structure~\cite{wang2018non,vaswani2017attention,sutskever2012imagenett}. 
This spatial/temporal structure is used to build a tensor such that the order of the feature vectors in the tensor exhibits the inherent spatial/temporal structure of the data.  For instance, the nearby pixels in an image are spatially correlated or the nearby embedding vectors (in the matrix which represents a sentence) are temporally correlated. The spatial/temporal ``order'' of the data is effectively used by the existing neural networks. For instance, the Convolutions Neural Network (CNN) leverages the local coherence of the pixels  to perform computationally efficient local feature extraction. In addition, the spatial order of the feature vectors are used to efficiently down-sample (pool) the extracted local features and perform  feature aggregation. 
A graph can also represent the order of the feature vectors. The sequence structure  in text data and the 2D structure in an image  can be viewed as special instances of graphs. 
If each node represents a feature vector, the structure of the graph exhibits a topological order of the feature vectors. The structure of the graph can be utilized~to perform local feature aggregation and  non-local feature aggregation via graph down-sampling~\cite{ying2018hierarchical} or spectral graph convolution~\cite{defferrard2016convolutional}. 

However, in many applications  there is not a fixed structure which can describe the given data. For instance, in some applications although each data in the dataset is represented using a graph, the graphs are not necessarily the same~\cite{KKMMN2016}. 
 For instance, the graphs which represent different toxic molecules are different  and a toxic molecule detector should be able to handle graphs with different structures. In this paper, we focus on the graph classification task, where despite the categorical similarities, the graphs within a class might be substantially different. Since a fixed structure cannot describe all the data samples, typically deep learning based graph classification approaches  are limited  to extracting the local features and aggregating all the local features using an indiscriminate aggregation function such as the element-wise max/mean function~\cite{duvenaud2015convolutional,bronstein2017geometric,atwood2016diffusion}. If one aims to use the graph structure to perform non-local feature aggregation, the structure of each graph should be analyzed separately. For instance,  
 in order to perform hierarchical feature extraction, 
 a graph clustering method can be used on each  graph individually~\cite{simonovsky2017dynamic,fey2018splinecnn,ying2018hierarchical}. This strategy however can be computationally prohibitive.  The complexity of the graph clustering/pooling algorithms mostly scale with $n^{2}$, where $n$ denotes the number of nodes.

In this paper, a new approach is presented using which the extracted local feature vectors  are transformed to  a latent representation  which resides on a Latent Fixed Data Structure (LFDS). The different input graphs are transformed to different signals over the LFDS. Several LFDSs are introduced, including predefined structures and data driven structures.  Our contributions can be summarized as follows:

\begin{itemize}
    \item We introduce an end-to-end differentiable~global feature aggregation  approach~in~which~a~set~of~unordered feature vectors are sorted/distributed~on~an~LFDS~and~the structure of the LFDS is leveraged to employ CNN or GNN to aggregate the representation~of~the~data~on~the~LFDS.\\ 
    
    \vspace{-0.08in}
    
    \item Several predefined LFDSs including 3D tensor (image), loop, and sequence are introduced. In addition, we propose to learn the structure of the LFDS in a data driven way and the presented LFDSs are used to design multiple new global feature aggregation methods. While the computational complexity of the proposed methods are linear with the order of input graphs, they achieve competitive or better results. 
    
\end{itemize}

\noindent\textbf{Notation:}
We use bold-face upper-case letters to denote matrices and bold-face lower-case letters to denote vectors. Given a matrix $\bX$, $\bx_{i}$ denotes the $i^{\text{th}}$ row of $\bX$. The inner product of two row or column vectors $\bx$, $\by$ is denoted by  $\big<\bx,\by\big>$.
A graph with $n$ nodes is represented by two matrices $\bA \in \mathbb{R}^{n \times n}$ and $\bX \in \mathbb{R}^{n \times d}$, where $\bA$ is the adjacency matrix, and $\bX$ is the  matrix of  feature vectors of the nodes.    The operation $\bA \Leftarrow \bB$ means that the content of $\bA$ is set equal to the content of $\bB$.  

\section{Related work} 
In recent years, there has been
a surge of interest in developing deep network architectures which can work with graphs~\cite{DBLP:conf/asunam/Shrivastava014,kipf2016semi,niepert2016learning,kipf2018neural,hamilton2017representation,fout2017protein,gilmer2017neural,tixier2018graph,simonovsky2017dynamic,bruna2013spectral,bronstein2017geometric,duvenaud2015convolutional,li2015gated}.
The main trend is to adopt the structure of CNNs and  most of the existing graph convolution layers can be loosely divided into two main subsets: the spatial convolution layers and the spectral convolution layers. 
The spectral methods are based on the generalization of spectral filtering in signal processing to the signals supported over graphs. The signal residing on the graph is transformed into the spectral domain using a graph
Fourier transform which is then filtered using a learnable filter's weight vector. Finally, the signal is transformed back to the nodal domain via inverse Fourier transform \textcolor{black}{~\cite{bruna2013spectral,defferrard2016convolutional,henaff2015deep,levie2017cayleynets}}. Upon defining a Fourier basis  $\bU \in \mathbb{R}^{n\times n}$, a spectral convolution layer can be written as
\begin{eqnarray}
\bX \Leftarrow f \left( \bU \mathbf{\Theta} \bU^T \bX \right) \:,
\end{eqnarray}
where $\mathbf{\Theta}$ is a diagonal matrix whose diagonal values are the parameters of the filter and $f(\cdot)$ is an element-wise non-linear function. The basis $\bU$ can be eigenvectors of the Laplacian matrix or the graph adjacency or their normalized versions.
The approach proposed in~\cite{bruna2013spectral} promotes the spatial locality of  the spectral filters  by designing smooth spectral filters. 
The major drawback of the spectral based methods is the non-generalizability to data residing over multiple graphs which is due to the dependency on the graph basis of Laplacian. 

In contrast to spectral graph convolution, the spatial graph convolution performs the convolution directly in the nodal domain and can be generalized across graphs. Similar to the convolution layer of a CNN, the spatial convolution layer aggregates the feature vector of each node with its neighbouring nodes~\cite{niepert2016learning,zhang2018end,nguyen2018learning,simonovsky2017dynamic,schlichtkrull2018modeling,velivckovic2017graph}. If the sum function is used to aggregate the local feature vectors, a simple spatial convolution layer can be written as  
\begin{eqnarray}
\bX \Leftarrow  \bA f \left( \bX \Phi_2 \right) + f \left(  \bX \Phi_1 \right)\:.
\label{eq:spatial}
\end{eqnarray}
The weight matrix $\mathbf{\Phi_1}$ transforms the feature vector of the given node and  $\mathbf{\Phi_2}$ transforms the feature vectors of the neighbouring nodes.

In order to  obtain a global representation of the graph, the local feature vectors obtained by the convolution layers should be aggregated into a final feature vector. The element-wise max/mean function is invariant to any permutation of its input and  is used widely to  aggregate all the local feature vectors.  Inspired by the hierarchical feature extraction in the CNNs,  tools have been developed to perform non-local feature aggregation~\cite{duvenaud2015convolutional,gao2019graph,lee2019self,zhang2018end,ying2018hierarchical,rahmani_geo,Proc:Rahmani_ICDM20,Proc:Rahmani_ICDM20,lee2019self,zhang2019hierarchical}. In~\cite{zhang2018end}, the nodes are ranked and the ranking is used to build a sequence of nodes using a subset of the nodes. Subsequently, a 1-dimensional CNN is applied to the sequence of the nodes to perform non-local feature aggregation. However, the way that~\cite{zhang2018end} builds the sequence of the nodes  is not data driven and there is no mechanism to ensure that relevant feature vectors are placed close to each other in the sequence.  The differentiable graph down-sampling method proposed in~\cite{ying2018hierarchical} learns an assignment matrix to downsize the input graph to a latent graph and the adjacency matrix of the latent graph is computed via down-sizing the adjacency matrix of the input graph using the computed assignment matrix.
The approach proposed in~\cite{gao2019graph,lee2019self} utilized a learnable query to select a set of nodes and downsize the input graph.  

\section{Global Feature Aggregation Using Latent Fixed Structures}
\begin{figure*}[t]
\centering

{\hspace{-0.3in}
		\includegraphics[width=0.7\textwidth]{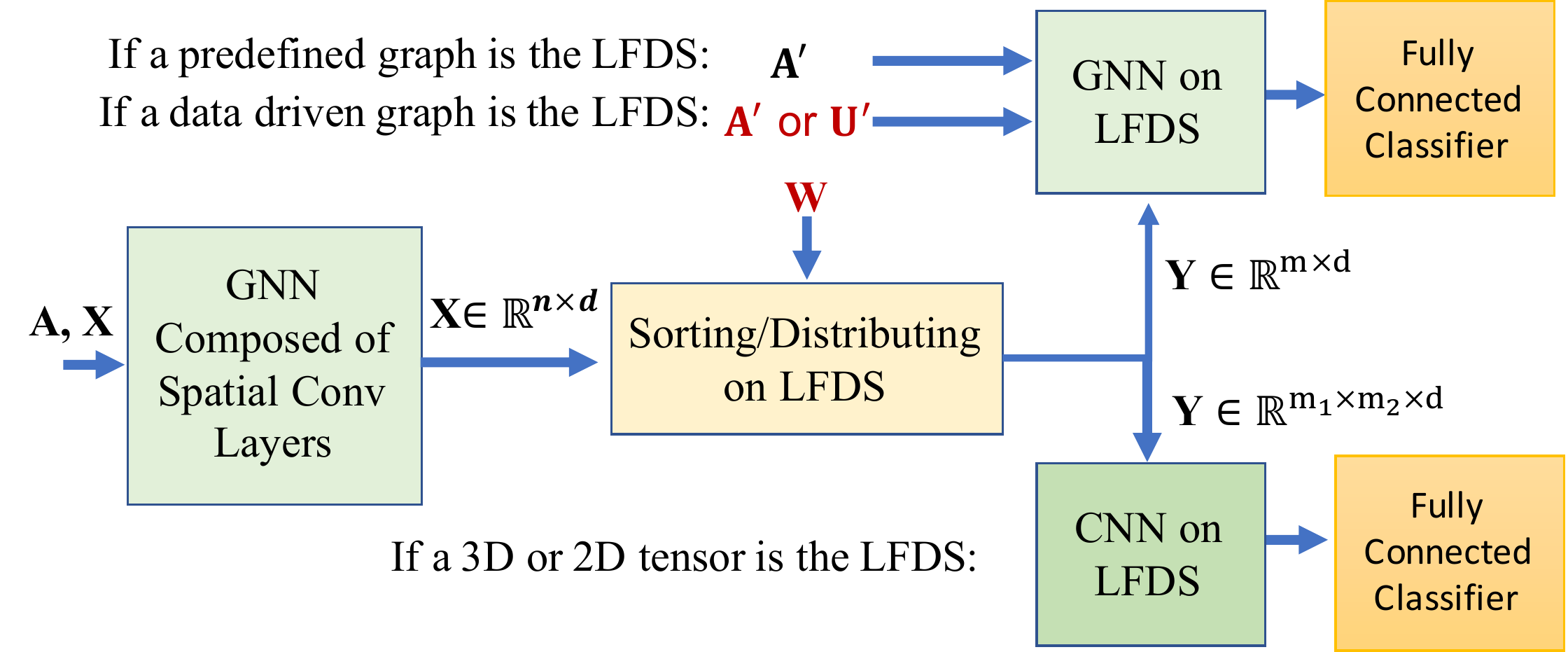}
		}

	\caption{This figures demonstrates the pipeline of the proposed approach with different LFDSs. The matrices/tensors in red are parameters of the neural network which are learned along with the other parameters of the neural network during the training process. The final classifier is a fully connected neural network which estimates the label of the input graph based on the representation vector obtained by   the latent GNN/CNN. }
	\label{f:figure_G2I}
\end{figure*}

\begin{figure*}[t]
\centering
{\hspace{-0.3in}
		\includegraphics[width=0.7\textwidth]{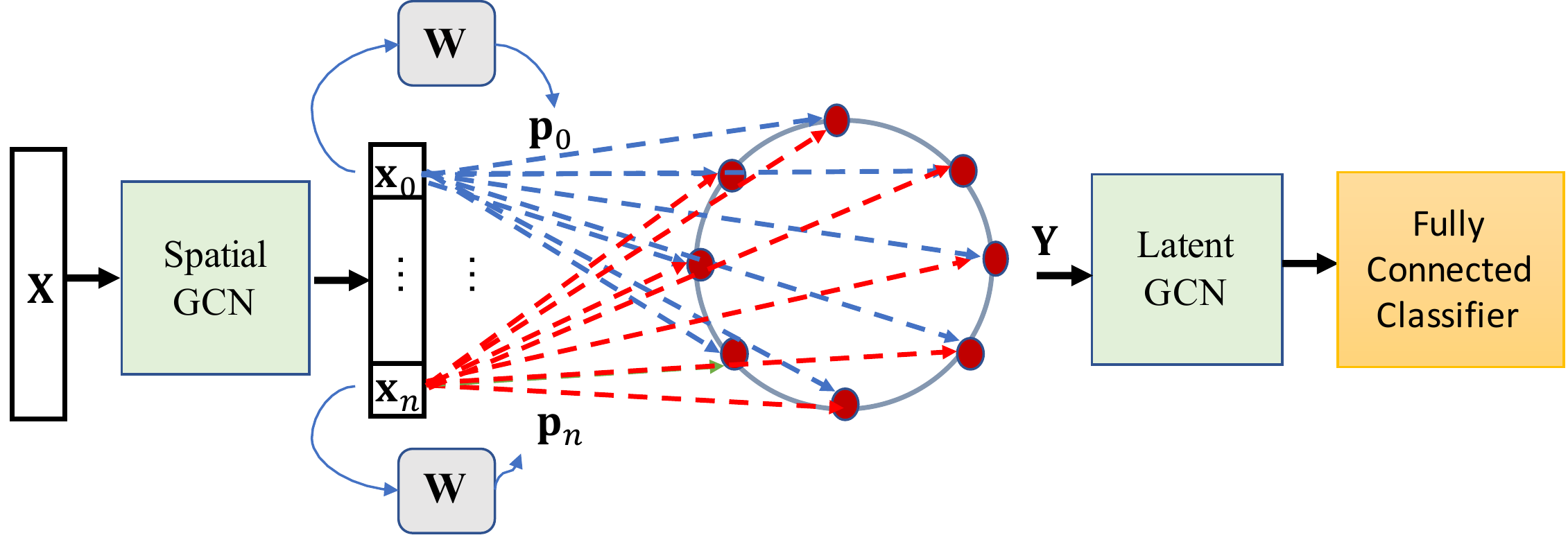}
		}

	\caption{This figure shows an example of the proposed approach in which a loop with 8 nodes is utilized as the LFDS. Accordingly, in this architecture $\bY \in \mathbb{R}^{8 \times d}$ and $\bA^{'}\in \mathbb{R}^{8 \times 8}$ where $\bA^{'}$ describes the connections which form a loop. Since in this example the LFDS is not data driven, $\bA^{'}$ is defined as a fixed matrix (not a parameter of the neural network). }
	\label{f:loop_example}
\end{figure*}

Suppose matrix $\bX \in \mathbb{R}^{n \times d}$ contains the input (potentially unordered) feature vectors. 
In this paper, $\bX$ is the matrix of local feature vectors which we need to aggregate them into a final representation vector. 
In the proposed approach, we distribute the input feature vectors on an LFDS. The LFDS is chosen such that a secondary neural network can process the feature vectors on the LFDS and aggregate them into a final feature vector.  For instance, assume the LFDS is a 3D tensor (image) and suppose that we properly distribute or sort the rows of $\bX \in \mathbb{R}^{n \times d}$ into the $d$-dimensional rows of 3D tensor $\bY \in \mathbb{R}^{m_1 \times m_2 \times d}$, i.e., each row of $\bX$ is assigned to one or several pixels of $\bY$ (by proper sorting/distributing we mean that the sorting/distributing algorithm respects the structure of the LFDS and places relevant feature vectors close to each other). Subsequently, we can apply a CNN to the tensor $\bY \in \mathbb{R}^{m_1 \times m_2 \times d}$ to aggregate the projection of the feature vectors on the LFDS and obtain the final representation vector. Figure~\ref{f:figure_G2I} exhibits an overview of the presented approach and Figure~\ref{f:loop_example} shows the proposed approach when the LFDS is a loop-graph.   In the following sections, the details of the performed steps are explained. First, the predefined and the data driven LFDSs are introduced. Next, the sorting/distributing method is presented. 


\subsection{Latent Fixed Data Structures (LDFSs)}
\label{sec:lfds}
In the recent years, significant progress has been made on the design of CNNs and GNNs. \emph{Therefore, we choose our LFDSs such that we can utilize a CNN or a GNN to process the data on the LFDS. }  We propose two sets of LFDSs: predefined LFDSs and data driven LFDSs. The structure of a predefined LFDS is chosen priorly. For instance, if the LFDS is a graph, all the connections between different nodes are chosen by the human designer. In contrast, with the data driven LFDSs, some details of the LFDS are learned in a data driven way. For instance, if the LFDS is a data driven graph, the adjacency matrix of the LFDS is defined as a parameter of the neural network to be learned during the training process. 

\subsubsection{Predefined Structures}
The structure of the predefined LFDSs is fixed and the neural network learns to perform feature aggregation over it. In this paper, we propose the following predefined LFDSs.

\vspace{0.08in}

\noindent
$\bullet$ \textbf{3D Tensor (Image): } 2D CNN is a powerful neural architecture which uses a sequence of  local feature aggregation layers and pooling layers to perform hierarchical feature extraction. If we choose the LFDS a 3D tensor, a 2D CNN can be utilized to process the data on the LFDS. Suppose the size of the 3D tensor is $m_1 \times m_2 \times d$ where $m_1\: m_2$ is the number of pixels of the corresponding image and $d$ is the~length~of~the feature vectors. 
Define tensor $\bY \in \mathbb{R}^{m_1 \times m_2 \times d}$ as the projection of $\bX$ onto the LFDS (next section presents methods which distribute/project  input feature vectors $\bX$ onto~the LFDS and obtain $\bY$).
The CNN is applied to tensor $\bY$ to aggregate~its feature vectors into a final representation vector  in a data~driven~way. 

\vspace{0.08in}

\noindent
$\bullet$ \textbf{Array: } This LFDS paves the way to use a 1D CNN to process the data on the LFDS whose computation complexity and memory requirement are less than a 2D CNN. Define $\bY \in \mathbb{R}^{m \times d}$ where $m$ is the length of the array corresponding to the LFDS. The 1D CNN is applied to $\bY$ to aggregate its feature vector into a final feature vector. 

\vspace{0.05in}

\noindent
$\bullet$ \textbf{Sequence-Graph:} 
The structure of this LFDS is similar to  Array   but we consider the array as a sequence graph which means that each row of $\bY$ is corresponding to one node of a sequence graph and a GNN (instead of 1D CNN) is used to process $\bY$. Define $\bA^{'} \in \mathbb{R}^{m \times m}$ as the adjacency matrix of the LFDS. The spatial convolution layer of the GNN which is applied to $\bY$ can be written~as 
\begin{eqnarray}
\bY \Leftarrow  \bA^{'} f \left( \bY \Phi_2^{'} \right) + f \left(  \bX \Phi_1^{'} \right)\:,
\label{eq:spatial2}
\end{eqnarray}
where $\Phi_1^{'}$ and $\Phi_1^{'}$ are the parameters of the convolution layer. 
The matrix $\bA^{'}$ is given to the neural network as a fixed matrix. As a simple example, if $m = 3$ then $\bA^{'}=[[0,1,0],[1,0,1],[0,1,0]]$ (in python notation).

\vspace{0.05in}

\noindent
$\bullet$ \textbf{Loop-Graph:} 
This LFDS is similar to  Sequence Graph  but it is assumed that the nodes of the latent graph form a loop. Similarly, with this LFDS, a latent GNN is used to perform feature aggregation on $\bY$ and (\ref{eq:spatial2}) describes the spatial graph convolution which is applied to $\bY$. Matrix $\bA^{'} \in \mathbb{R}^{m \times m}$ represents the adjacency matrix of a loop  and it is given to the neural network as a fixed matrix. 

\vspace{0.05in}

\subsubsection{Data Driven Structures}
In this section, we introduce an LFDS whose structure is learned in a data driven way. Graph is a flexible and descriptive structure. One important feature of a graph is that all the information about its structure lies in  its adjacency matrix. This is a desired feature because we can simply define the adjacency matrix of the latent graph as a parameter of the neural network to be  learned along with the other parameters of the neural network during the training process. \emph{A significant feature of the LFDS is that its structure is fixed for all the input graphs and both spatial convolution and spectral convolution can be utilized to perform feature aggregation over the latent graph.} We present two methods based on the choice of the convolution layer. 

\vspace{0.05in}

\noindent
$\bullet$ \textbf{Parametric Graph with Spatial Convolution:} 
If spatial convolution is used in the latent GNN (which is applied to $\bY$), then we can define $\bA^{'} \in \mathbb{R}^{m \times m}$ used in (\ref{eq:spatial2}) as a parameter of the neural network. Accordingly, the neural network learns the connection between the nodes of the LFDS in a data driven way. However, we have to ensure that 
the learned adjacency matrix satisfies the properties of an adjacency matrix for  undirected graphs, namely symmetry, nonnegativity of the entries and zeros entries on the diagonal. We can achieve these properties by imposing the appropriate structure on $\bA^{'}$. In particular, instead of the adjacency  matrix $\bA^{'}$, we define  $\bB \in \mathbb{R}^{m \times m}$ as the neural network parameter which is used to construct $\bA^{'}$ as $$\bA^{'} = \sigma(\mathbf{B} + \mathbf{B}^{T}) - \text{Diag}\Big(\text{diag}\big(\sigma(\mathbf{B} + \mathbf{B}^{T})\big)\Big)\:,$$ where  $\sigma(.)$ denotes the sigmoid activation function. Accordingly, $\bA'$ satisfies the required conditions by construction.

\vspace{0.05in}

\noindent
$\bullet$ \textbf{Parametric Graph with Spectral Convolutions:} 
If spectral convolution is used to perform feature aggregation on the latent graph, we need to compute the Fourier Basis of the latent graph. A simple solution to avoid computing the Fourier Basis is to define the Fourier basis itself as a parameter of the neural network. Accordingly, if  spectral convolution layers are used in the latent GNN which is applied to $\bY$, we no longer need to define $\bA^{'}$ as a parameter of the neural network and we directly define $\bU^{'} \in \mathbb{R}^{m \times m}$ as the Fourier Basis of the latent graph a parameter of the neural network. Accordingly, the spectral convolution applied to the latent graph can be written as 
\begin{eqnarray}
\bY \Leftarrow f \left( \bU^{'} \mathbf{\Theta}^{'} {\bU^{'}}^T \bY \right) \:,
\end{eqnarray}
where both $\mathbf{\Theta}^{'}$ and $\bU^{'}$ are the parameters of the neural network. However, we need to ensure that the learned $\bU^{'}$ is orthonormal. Accordingly, if Parametric Graph with Spectral Convolutions is used as the LFDS, we add $\lambda \lVert \bU'^{T} \bU' - \mathbf{I} \rVert_{F}^{2}$ to the final cost function which is used to train the neural network where  $\lambda$ is a regularization coefficient. This regularizer encourages the optimizer to find $\calU^{'}$ on the manifold of orthonormal matrices. 


\subsection{Sorting/Distributing the Input Data on the LFDS}
In Section~\ref{sec:lfds}, it was assumed that there is an algorithm which  sorts/distributes the input feature vectors $\bX$ on the LFDS and form $\bY$ (the data on the LFDS). In this section, an algorithm is presented which is invariant to the permutation of the input feature vectors. 
In this algorithm, we define  $\bW \in \mathbb{R}^{m \times d}$ as a parameter of the neural network where $m$ is the number of nodes/elements of the LFDS.  Each row of $\bW$  corresponds to one node/element of the LFDS and $\bW$ is used for sorting/distributing the input feature vectors on the LFDS. Note that if the LFDS is a 3D tensor as described in Section~\ref{sec:lfds}, then $\bW \in \mathbb{R}^{m_1 \times m_2 \times d}$. 

\vspace{0.05in}

\begin{remark}
Without loss of generality, in this section it is assumed that the LFDS is a graph which means that $\bY \in \mathbb{R}^{m \times d}$ and the LFDS is composed of $m$ nodes. The presented algorithms are similarly applicable to all the other LFDSs. 
Each row of matrix/tensor $\bW$ can be considered as a learnable query and learnable queries have been used in the previous works \cite{ying2018hierarchical} for feature aggregation. In this paper, we impose a (predefined or data driven) structure on the queries using the LFDS and the LFDS is utilized to aggregate the feature vectors obtained using the queries. 
\end{remark}

\vspace{0.08in}
\noindent\textbf{Projecting the Data on the LFDS. }
Each row of $\bW$ represents one node of the LFDS. We use the rows of $\bW$ to measure the relevance of each input feature vector with the nodes of the LFDS. Specifically, define vector $\bp_i \in \mathbb{R}^{m}$ corresponding to $\bx_i$ (the $i^{th}$ row of $\bX$) as 
\begin{eqnarray}
\bp_i (k)  = \frac{ \exp \Big( \big< \bx_i, \bw_k \big> \Big)}{\sum_{u} \exp \Big( \big< \bx_i, \bw_u \big> \Big) } \:,
\label{eq:distrgh}
\end{eqnarray}
where $\bw_k$ is the $k^{th}$ row of $\bW$, $\bp_i (k)$ is the $k^{th}$ element of vector $\bp_i$, and $<\cdot,\cdot>$ measures the similarity between its component where in this paper we use inner-product. Vector $\bp_i$ represents the similarity between $\bx_i$ and the nodes of the LFDS. Accordingly, we utilize $\{ \bp_i \}_{i=1}^{n}$ to distribute $\{ \bx_i \}_{i=1}^n$ on the LFDS by defining matrices $\{ \bY_i \in \mathbb{R}^{m \times d} \}_{i=1}^n$ as
\begin{eqnarray}
\bY_i = \bp_i \otimes \bx_i 
\end{eqnarray}
which are used to compute $\bY$ as 
\begin{eqnarray}
\bY = \sum_{i=1}^n \bY_i \:.  
\label{eq:Yfinal}
\end{eqnarray}

\vspace{0.05in}
In the proposed approach,
we used the $d$-dimensional rows of $\bW$ as representation vectors for the elements of the LFDS. The tensor/matrix $\bW$ is learned during the training process along with other parameters of the neural network. Since the $d$-dimensional rows of $\bW$ represent the elements of the LFDS, we expect them to follow the structure of the LFDS. For instance, suppose the LFDS is an image and $\bw_i$ and $\bw_j$ are corresponding to the $i^{th}$ pixel and the $j^{th}$ pixel of the LFDS, respectively. We expect $\bw_i$ and $\bw_j$ to be coherent with each other in feature space if the  the $i^{th}$ pixel and the $j^{th}$ pixel are close and vice versa. 
The optimizer learns the $d$-dimensional feature vectors of $\bW$ such that 
the GNN/CNN can successfully perform feature aggregation on the LFDS and  this happens when the coherency between the $d$-dimensional feature vectors of $\bW$ follows the structure of the LFDS. 
In other word, the presence of the latent GNN/CNN automatically bounds the structure of the LFDS with the distribution of the rows of $\bW$.

\vspace{0.02in}
\begin{remark}
To  ensure that the neural network uses all the elements of the LFDS and does not end up using a few elements, we employ an element dropout technique during the training process, i.e., in each training iteration, a random subset of the $d$-dimensional rows of $\bY$ are set to zero. 
\label{remark:drop}
\end{remark}

\vspace{0.02in}

\begin{remark}
The proposed approach utilizes the feature vectors in  $\bW$ to project the information in $\bX$ onto the elements of the LFDS. There are other global feature aggregation methods such as global max-pooling which are
computationally efficient aggregation methods and we can utilize them inside the proposed  method.
For instance, suppose we want to include the information obtained by 
 global max pooling in the proposed  method. Define vector $\bx_{\max} \in \mathbb{R}^{d}$ as the element-wise max pooling of the rows of $\bX$. Each row of $\bY$ is a representation vector for the input data $\bX$. Thus, we can simply aggregate $\bx_{\max}$ with each row of $\bY$. Using this technique, we can include the information obtained by any other global feature aggregation method in $\bY$. In the presented experiments, we aggregated $\bx_{\max}$ with all the rows of $\bY$. Specifically, we updated each row of $\bY$ as $\by_i \Leftarrow a_1 \by_i + a_2 \bx_{\max}$ where $a_1$ and $a_2$ are two positive coefficients which were defined as the parameters of the neural network to let the neural network learn the best linear combination of them. 
\end{remark}

 \begin{table*}[h]
\begin{center}{
\caption{Classification accuracy with different non-local feature aggregation methods. 
}
\begin{tabular}{l |lcccc}
\hline\hline
  &     PC Graphs  &  PROTEINS  & DD & ENZYM & SYNT  \\
\hline
  Max Pooling &     49.2 $\pm$ 9.8  &  80.5 $\pm$ 3.1 & 83.3 $\pm $ 3.2 & 69.2 $\pm$ 4.5 & 71.2  $\pm $ 2.5 \\
 Graph-Loop  &    \textbf{55.2} $\pm$ 2.2  &  80.9 $\pm$ 2.9  & 83.5 $\pm$ 3.5 & \textbf{74.3} $\pm$ 3.8 & 72.4 $\pm$ 3.4   \\
 Graph-Sequence  &  53.7 $\pm$ 10.9  &  80.9 $\pm$ 3.0  & 83.7 $\pm$ 2.9 & 72.0 $\pm$ 4.0 & 72.2 $\pm$ 2.9  \\
 Data Driven  (spatial) &     51.4 $\pm$ 10.6  &  80.4 $\pm$ 3.3  & \textbf{84.6} $\pm$ 2.8 & 72.0 $\pm$ 4.0 & 72.5 $\pm$ 4.4   \\
  Data Driven  (spectral) &     48.0 $\pm$ 7.8 &   80.7 $\pm$ 3.2  & 83.4 $\pm$ 2.8 & 71.7 $\pm$ 4.1 & \textbf{73.0} $\pm$ 3.1   \\
 3D Tensor (image) &     \textbf{55.0} $\pm$ 1.9  &  \textbf{81.4} $\pm$  3.4  & \textbf{84.4} $\pm$ 2.8 & 71.0 $\pm$ 5.3 & \textbf{72.7} $\pm$ 1.3   \\
2D Tensor (array) &  52.1 $\pm$ 10.6      &   80.6 $\pm$ 3.5  & 83.7 $\pm$ 3.8 &  71.8 $\pm$ 4.0& \textbf{73.0} $\pm$ 2.4   \\
 Sort-Pooling &     49.7 $\pm$ 2.3  &  80.5 $\pm$ 4.5 & 82.4 $\pm$ 2.5 & 63.2 $\pm$ 3.7 & 67.7 $\pm$ 3.0  \\
 Diff-Pool &     42.5 $\pm$ 16.7  &  80.4 $\pm$ 3.4  & \textbf{84.4} $\pm$ 2.7  & 70.3 $\pm$ 5.2 & {71.8}  $\pm$ 3.3  \\
 Rank-PooL &     39.8 $\pm$ 11.8  & {80.7} $\pm$ 3.1 & 83.7 $\pm$  3.8  & 67.5 $\pm$ 7.7 & 71.5 $\pm$ 2.8  \\
\hline\hline
\end{tabular}
}
\end{center}
\label{tab:real_data_kernels}\vspace{-0.2in}
\end{table*}

\section{Numerical Experiments}

The proposed methods are compared with some of the existing  graph feature aggregation methods including DiffPool~\cite{ying2018hierarchical}, Node-Sort~\cite{zhang2018end}, and Rank-PooL~\cite{gao2019graph,lee2019self}. 
Following the conventional settings, we perform $10$-fold cross validation: $9$ folds for training and 1 fold~for~testing. 
%
We study the performance of different feature aggregation methods with 5 datasets. 
The utilized datasets include  4 benchmark graph classification datasets and 
we refer the reader to~\cite{KKMMN2016} and the references therein for more information about them. We also created a graph classification dataset using point cloud data. In this new dataset (PC Graphs), the GNNs are trained to classify point clouds solely based on their nearest neighbour graphs. The location of the points in not provided and the GNN is required to distinguish point clouds~in~different~classes~based~on the differences between their nearest neighbour graphs. The dataset is composed of nearly 4000 graphs and the graphs~form~8 classes. Each graph  corresponds 
to the nearest neighbour graph of a point cloud which is composed of 150 points. We use the points clouds in the ShapeNet dataset~\cite{chang2015shapenet} and the chosen shapes are: Table, Airplane, Car, Guitar, Knife, Lamp, Chair, and Laptop.  
In the following, we describe the data preprocessing steps and our architecture~of~the~neural~networks. 

\vspace{0.08in}

\noindent
\textbf{The input to the neural networks.}  We follow the approach presented in~\cite{rahmani_geo,Proc:Rahmani_ICDM20} to leverage both the node labels/attributes and the node embedding vectors. It was shown in~\cite{rahmani_geo,Proc:Rahmani_ICDM20} that embedding vectors make the neural network more aware of the topological structure of the graph. The Deep-Walk graph embedding method~\cite{perozzi2014deepwalk} is used to embed the graphs and the dimension of the embedding vectors are set to $12$. If $s$ is the size of the graph, the length of the random walks is determined as $\max \left( 4 , \min(s/10, 10)\right)$. 

\vspace{0.05in}

\noindent
\textbf{The structure of the basis neural network.}
In the presented experiments, the GNN equipped with spatial representation proposed in~\cite{rahmani_geo,Proc:Rahmani_ICDM20} is used to extract the matrix of local feature vectors $\bX$. The implemented shared neural network which extracts the local feature vectors is composed of three spatial convolution layers and each convolution layer contains two weight matrices as in (\ref{eq:spatial}).  The dimensionality of the output of all the convolution layers  is equal to 64. Each convolution layer is equipped with batch-normalization~\cite{ioffe2015batch} and ReLu is used as the element-wise nonlinear function. 
 The output of the last convolution layers is used as the input of  the non-local feature aggregation methods (except the Sort-Node method for which the concatenation of the outputs of  all the three convolution layers was used).  
 
 \vspace{0.05in}
 
 \noindent
 \textbf{The final representation vector of the input graph.}
 Define $\bX \in \mathbb{R}^{n \times 192}$ as the concatenation of all the spatial convolution layers of the basis neural network and define $\bx \in \mathbb{R}^{1\times 192}$ as the element-wise max-pooling of the rows of $\bX$. In addition, define $\bY$ as the output of the non-local aggregation method and define $\by$ as the element-wise max-pooling of the rows of $\bY$. The final representation of the graph for all the methods (except Sort-Node~\cite{zhang2018end}) is obtained as the concatenation of vectors $\bx$ and  $\by$. For the Sort-Node method~\cite{zhang2018end}, the output~of the pooling method is used as the representation of the~graph. 

\vspace{0.05in}

\noindent
\textbf{The structure of the proposed methods.}
In the following, we describe the architecture of the proposed approach with 2 LFDSs. The implemented architectures with the other LFDSs are similar.

\vspace{0.05in}
\noindent$\bullet$	Proposed method with 3D tensor as the LFDS: 
We define $\bY$ equal to a tensor with 64 feature vectors, i.e., $\bW \in \mathbb{R}^{8 \times 8 \times d}$. Two CNN convolution layers are used to process $\bY \in \mathbb{R}^{d \times 8 \times 8}$ and the size of the convolution kernel in both the convolution layers is $3\times 3$. Define $\by_1 \in \mathbb{R}^{1 \times d}$ as the aggregation (using max-pooling) of all the feature vectors of $\bY$ after the first convolution layer and define $\by_2 \in \mathbb{R}^{1 \times d}$ as the  aggregation (using max-pooling) of all the feature vectors of $\bY$ after the second convolution layer. The vector $\by$ is defined as the concatenation of $\by_1$ and $\by_2$. The final representation of graph is defined as the concatenation of $\by$ and $\bx$. The dimensionality of the final representation vector is 320 (192 + 128 = 320). 

\vspace{0.05in}

\noindent $\bullet$ Proposed method with graph as the LFDS: First suppose that the latent GCN, which is used to process $\bY$, is composed of spatial convolution layers. 
The latent GCN is composed of two spatial convolution layers and the functionality of each convolution layer can be written as (\ref{eq:spatial2}) where $f(\cdot)$ is the Relu function followed by Batch-normalization. If the LFDS is a data driven graph, $\bA^{'}$ is defined as a parameter of the neural network. Similar to the proposed method with 3D tensor, the output of both convolution layers are used to build $\by$ and the last representation vector is built similarly. 

\vspace{0.05in}

\noindent If the latent GCN is made of spectral graph convolution layers,
the functionality of each implemented spectral convolution layer can be written as
$$
\bY \Leftarrow f \left( \bU^{'} \mathbf{\Theta} {\bU^{'}}^T \bY {\Phi_2}^{'} \right) + f \left( \bY {\Phi_1}^{'} \right)\:,
$$
where $\bU^{'} \in \mathbb{R}^{m\times m}$ is the Fourier basis of the latent graph, the diagonal matrix $\mathbf{\Theta}^{'} \in \mathbb{R}^{m\times m} $ is the weight matrix of the spectral convolution layer,  ${\Phi_2}^{'}$ and ${\Phi_1}^{'}$ perform feature transformation, and  $f(\cdot)$ represents Relu function followed by Batch-normalization.  

\vspace{0.08in}

\noindent
\textbf{The final classifier:} A 3-layer fully connected neural network  transforms the final vector representation of the graph   to a $c$-dimensional vector where $c$ is the number of classes. 

\vspace{0.05in}
\noindent\textbf{Training:}
In order to avoid over-fitting, we utilize different dropout techniques: dropout on the final fully connected layers with probability 0.5, dropping out each node feature vectors with probability 0.2, and random dropout of the elements of the LFDS with probability 0.4 (see Remark~\ref{remark:drop}). 
The cross entropy function is used as the classification loss and all the neural networks are similarly trained using the Adam optimizer~\cite{kingma2014adam}.
The learning rate is initialized at $0.005$ and is reduced to $0.0001$ during the training process. 

\vspace{0.08in}
\noindent\textbf{The baselines: }

\vspace{0.05in}
\noindent
$\bullet$ Diff-Pool~\cite{ying2018hierarchical}: This method was implemented according to the instructions in~\cite{ying2018hierarchical}. Two spatial convolution layers were placed after the down-sizing step. The final representation of the graph was obtained similar to the procedure used in the proposed approach.  

\vspace{0.05in}
\noindent
$\bullet$ Rank-PooL~\cite{gao2019graph,lee2019self}:  Similar to the other pooling methods, one down-sampling layer was used and two spatial convolution layers were implemented after the down-sizing step. The final representation of the graph was obtained similar to the procedure used in the proposed method. 

\vspace{0.05in}
\noindent
$\bullet$ Sort-Node~\cite{zhang2018end}: Similar to the implementation described in~\cite{zhang2018end} and its corresponding code,  $30$ nodes were sampled to construct the ordered sequence of the nodes.

\newpage
Table~1 demonstrates the  classification accuracy for all the datasets.
One  observation is that on most of the datasets, the proposed methods outperform the simple GNN. The main reason for achieving a higher performance is that the  LFDS used in the proposed methods paves the way for the neural network to aggregate the extracted local feature vectors in a data driven way. In addition, on most of the datasets, the presented methods outperform the Sort-Node method proposed in~\cite{zhang2018end} and the main reason is that the way Sort-Node sorts the nodes in the one dimensional array does not necessarily put relevant nodes close to each other. In contrast, in the proposed approach, the extracted local feature vectors are ordered/distributed on the LFDS using the data driven  feature vectors $\bW$ and the presence of the latent GNN/CNN bounds $\bW$ with the structure of the LFDS.

\section{Future Works}
In the proposed approach, each row of $\bW$ corresponds to an element/node of the LFDS and the similarly between the rows of $\bW$ and the rows of $\bX$ is used to compute $\bY$. One might critique that 
a fixed set of feature vectors might not be diverse enough to work for all the graphs in the dataset and the the rows of $\bW$ should change conditioned on the input data.   
A possible extension to the proposed approach is an algorithm in which $\bW$ is generated by a side neural network whose input is   the given graph. For instance, suppose $\bW \in \mathbb{R}^{m\times d}$, define $\bx_{\max}$ as the max-pooling of the rows of $\bX$, and assume $h(\bx_{\max})$ is a function which maps $\bx_{\max}$ to a $m\times d$ dimensional matrix. The output of $h(\bx_{\max})$ can be used as the weight matrix $\bW$ to ensure that the utilized feature vectors is conditioned on the input graph. A possible scenario is to define a set of candidate weight matrices $\{ \bW_i \}_{i=1}^K$ such that the final weight matrix $\bW$ is computed as a weighted combination of them, i.e., $\bW = \sum_{i=1}^K \alpha(i) \bW_i $ where vector $\alpha \in \mathbb{R}^{K}$ is computed via transforming $\bx_{\max}$ using a fully connected neural network.

\section{Conclusion}
An end-to-end scalable framework for non-local feature aggregation over graphs and deep analysis of unordered data was proposed. The proposed approach projects the unordered feature vectors over a Latent Fixed Data Structure (LFDS) and the structure of the LFDS is used to aggregate the projected local feature vectors. It has been shown that the proposed approach can be used to design several new feature aggregation methods. We have introduced multiple structures for the LFDS including graph, tensor, and array. It was shown that the LFDS can be predefined and it can also to be  a learnable graph. If the LFDS is data driven (learnable graph), the adjacency matrix of the latent graph is defined as the parameter of the neural network. Overall, the presented experiments show that the proposed methods achieve  a competitive performance.

\bibliography{citations}
\bibliographystyle{plain}

\end{document}

%% file: preamble.tex
\usepackage[mathscr]{eucal}
\usepackage[cmex10]{amsmath}
\usepackage{epsfig,epsf,psfrag}
\usepackage{amssymb,amsmath,amsthm,amsfonts,latexsym}
\usepackage{amsmath,xcolor,bm}
\usepackage[caption=false]{subfig} 
\usepackage{fixltx2e}
\usepackage{array}
\usepackage{verbatim}
\usepackage{bm}
\usepackage{algorithmic, cite}
\usepackage{algorithm}
\usepackage{verbatim}
\usepackage{textcomp}
\usepackage{mathrsfs}
\usepackage{epstopdf}

\allowdisplaybreaks[3]

\newcommand{\calU}{\mathcal{U}}

\newcommand{\bA}{\mathbf{A}}

\newcommand{\bB}{\mathbf{B}}

\newcommand{\bp}{\mathbf{p}}

\newcommand{\bU}{\mathbf{U}}

\newcommand{\bw}{\mathbf{w}}
\newcommand{\bW}{\mathbf{W}}
\newcommand{\bx}{\mathbf{x}}
\newcommand{\bX}{\mathbf{X}}
\newcommand{\by}{\mathbf{y}}
\newcommand{\bY}{\mathbf{Y}}

\DeclareMathAlphabet{\mathbsf}{OT1}{cmss}{bx}{n}
\DeclareMathAlphabet{\mathssf}{OT1}{cmss}{m}{sl}

\DeclareSymbolFont{bsfletters}{OT1}{cmss}{bx}{n}  
\DeclareSymbolFont{ssfletters}{OT1}{cmss}{m}{n}
\DeclareMathSymbol{\bsfGamma}{0}{bsfletters}{'000}
\DeclareMathSymbol{\ssfGamma}{0}{ssfletters}{'000}
\DeclareMathSymbol{\bsfDelta}{0}{bsfletters}{'001}
\DeclareMathSymbol{\ssfDelta}{0}{ssfletters}{'001}
\DeclareMathSymbol{\bsfTheta}{0}{bsfletters}{'002}
\DeclareMathSymbol{\ssfTheta}{0}{ssfletters}{'002}
\DeclareMathSymbol{\bsfLambda}{0}{bsfletters}{'003}
\DeclareMathSymbol{\ssfLambda}{0}{ssfletters}{'003}
\DeclareMathSymbol{\bsfXi}{0}{bsfletters}{'004}
\DeclareMathSymbol{\ssfXi}{0}{ssfletters}{'004}
\DeclareMathSymbol{\bsfPi}{0}{bsfletters}{'005}
\DeclareMathSymbol{\ssfPi}{0}{ssfletters}{'005}
\DeclareMathSymbol{\bsfSigma}{0}{bsfletters}{'006}
\DeclareMathSymbol{\ssfSigma}{0}{ssfletters}{'006}
\DeclareMathSymbol{\bsfUpsilon}{0}{bsfletters}{'007}
\DeclareMathSymbol{\ssfUpsilon}{0}{ssfletters}{'007}
\DeclareMathSymbol{\bsfPhi}{0}{bsfletters}{'010}
\DeclareMathSymbol{\ssfPhi}{0}{ssfletters}{'010}
\DeclareMathSymbol{\bsfPsi}{0}{bsfletters}{'011}
\DeclareMathSymbol{\ssfPsi}{0}{ssfletters}{'011}
\DeclareMathSymbol{\bsfOmega}{0}{bsfletters}{'012}
\DeclareMathSymbol{\ssfOmega}{0}{ssfletters}{'012}

\newtheorem{remark}{Remark}

\newtheorem{data model}{Data Model}

\newcommand{\qednew}{\nobreak \ifvmode \relax \else
      \ifdim\lastskip<1.5em \hskip-\lastskip
      \hskip1.5em plus0em minus0.5em \fi \nobreak
      \vrule height0.75em width0.5em depth0.25em\fi}